\documentclass{IEEEtran}
\IEEEoverridecommandlockouts
\usepackage{cite}
\usepackage{threeparttable}
\usepackage{array}
\usepackage{amsmath,amssymb,amsfonts}
\usepackage{algorithmic}
\usepackage{graphicx}
\usepackage{textcomp}
\usepackage{makecell}
\usepackage{subfigure} 
\usepackage{xcolor}
\usepackage{algorithmic}
\usepackage[ruled]{algorithm2e}
\usepackage{multirow}
\usepackage{bm}
\usepackage{diagbox}
\usepackage{hyperref}
\hypersetup{
	colorlinks=true,
	linkcolor=red,
	filecolor=black,      
	urlcolor=black,
	citecolor=green,
}
\usepackage{booktabs}
\usepackage{soul}

\def\BibTeX{{\rm B\kern-.05em{\sc i\kern-.025em b}\kern-.08em
		T\kern-.1667em\lower.7ex\hbox{E}\kern-.125emX}}

\usepackage[top=0.45in,bottom=0.45in,left=0.4in,right=0.4in]{geometry} 

\makeatletter
\newcommand{\linebreakand}{%
\end{@IEEEauthorhalign}
\hfill\mbox{}\par
\mbox{}\hfill\begin{@IEEEauthorhalign}
}
\makeatother
\begin{document}
	\title{Learning Invariant Representation  via Contrastive Feature Alignment for Clutter Robust SAR Target Recognition
		\thanks{This work was partially supported by the 
	National Key Research and Development Program of China under Grant 
	2021YFB3100800, National Natural Science Foundation of China under Grant 
	61921001, 62022091, and the Changsha Outstanding Innovative Youth Training 
	Program under Grant kq2107002. \emph{(Corresponding author: Bo Peng.)}}
	}
	\author{\IEEEauthorblockN{Bowen Peng, Jianyue Xie, Bo Peng, and Li Liu, \emph{Senior Member, IEEE}}
		\thanks{Bowen Peng, Jianyue Xie, Bo Peng, and Li Liu are with the College of Electronic Science, National University of Defense Technology (NUDT), Changsha, 410073, China (\url{pbow16@nudt.edu.cn},  \url{xiejianyue0125@163.com},  \url{ppbbo@nudt.edu.cn}, \url{dreamliu2010@gmail.com}).}
	}

	\maketitle

	\begin{abstract}
		The deep neural networks (DNNs) have freed the synthetic aperture radar automatic target recognition (SAR ATR) from expertise-based feature designing and demonstrated superiority over conventional solutions. 
		There has been shown the unique deficiency of ground vehicle benchmarks in shapes of strong background correlation results in DNNs overfitting the clutter and being non-robust to unfamiliar surroundings. 
		However, the gap between fixed background model training and varying background application remains underexplored. Inspired by contrastive learning, this letter proposes a solution called Contrastive Feature Alignment (CFA) aiming to learn invariant representation for robust recognition. The proposed method contributes a mixed clutter variants generation strategy and a new inference branch equipped with channel-weighted mean square error (CWMSE) loss for invariant representation learning. 
		In specific, the generation strategy is delicately designed to better attract clutter-sensitive deviation in feature space. The CWMSE loss is further devised to better \textit{contrast} this deviation and \textit{align} the deep features activated by the original images and corresponding clutter variants. The proposed CFA combines both classification and CWMSE losses to train the model jointly, which allows for the progressive learning of invariant target representation.
		Extensive evaluations on the MSTAR dataset and six DNN models prove the effectiveness of our proposal. The results demonstrated that the CFA-trained models are capable of recognizing targets among unfamiliar surroundings that are not included in the dataset, and are robust to varying signal-to-clutter ratios.
	\end{abstract}
	
	
	\section{Introduction}
	\IEEEPARstart{A}{utomatic} target recognition (ATR) is a crucial and challenging task in synthetic aperture radar (SAR) image interpretation, and serves as the foundational technology in a variety of civil and military applications such as disaster rescue, reconnaissance, and surveillance, \textit{etc}. During the past decade, deep neural networks (DNNs), the data-driven and end-to-end training techniques, have significantly advanced SAR ATR with superiority in terms of effectiveness, efficiency, and many other aspects
	 \cite{kechagias2021automatic}.
	
	SAR ATR algorithms are expected to be highly generalizable and robust, in addition to being accurate, as the limited collection opportunities usually result in restricted training samples. However, the inspiring capability of DNN in automatic feature mining, which liberates SAR ATR from expertise-based feature extraction, may also trap itself in the data bias of the small-scale SAR dataset.	Consider a DNN for optical imagery analysis, which is trained with a hundred images of a tank parked on gravel ground (labeled as \textit{tank}) and another hundred of a truck parked on the meadow (labeled as \textit{truck}). Although the training images cover information about the targets from different viewing perspectives, the model is still likely to recognize any type of vehicle above gravel as a \textit{tank}, or \textit{vice versa}. 

	In this instance, the data bias manifests as a strong target-background correlation, and massive amounts of training data such as various kinds of vehicles among diverse surroundings, like the tens of thousands in the ImageNet dataset, could unlock the classifier to eliminate the influence of this bias to be well generalizable. On the contrary, the major benchmark dataset of SAR ATR, the MSTAR dataset, holds a deficiency of the target-clutter correlation in that the whole target images of a category are collected often in the same background \cite{schumacher2005non}. As a result, the clutters are also discriminative for recognition aside from the target and shadow, and allow high true positive ATR performance even with the absence of the target \cite{kechagias2021automatic}. 
	
	Although the target-clutter correlation in SAR data collection of the ground targets was demonstrated over years ago \cite{schumacher2004atr,schumacher2005non}, little attention has been paid to studying and alleviating its influence on the DNN-based ATR solutions. Zhou \textit{et al.} \cite{clutter} trained a DNN only with the labeled clutters of the MSTAR dataset and without any structural and scattering information carried by the target and shadow. Their results demonstrated that such a clutter-only-trained DNN can achieve a counter-intuitive recognition accuracy (over 40\%). Belloni \textit{et al.} \cite{AnylisiTAES} showed that for a trained classifier, the absence of clutter is nearly severe as that of the target, that is, the accuracy decreased from 98\% to 33\%, which to 27\% as the target got masked. These results indicated that the clutters within each class contain, indeed, shared features. Recently, Li \textit{et al.} \cite{weijie} showed that DNN overfitting on the clutters is partially contributed by the comparable signal-to-clutter ratios (SCR) across classes. Heiligers \textit{et al.} \cite{segclutter} suggested to train models only with the target segments for training and testing can mitigate the overfitting. However, the method suffers from the lack of precise model-based segmentation annotations for test images \cite{malmgren2015convolutional} that the ATR user is unlikely to get before the target is accurately categorized.

	\begin{figure*}[tbp]
		\centering
		\includegraphics[width=1\linewidth]{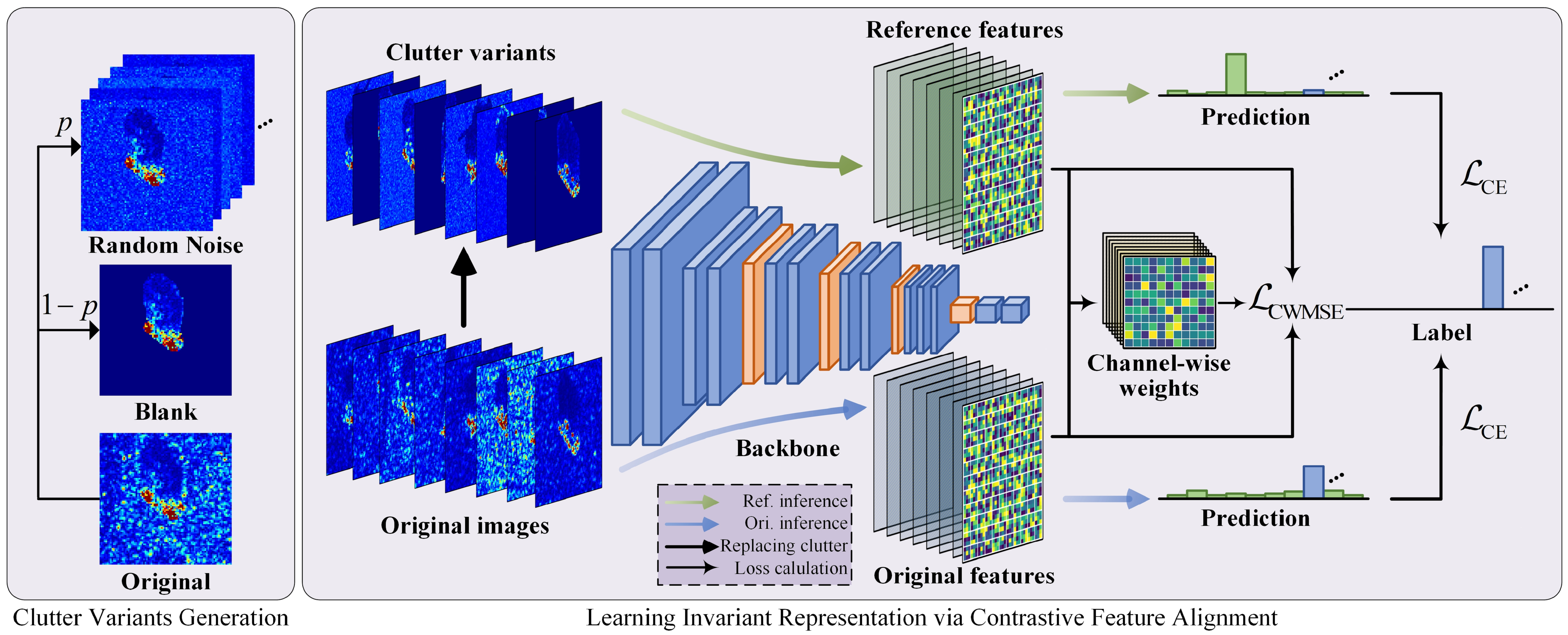}
			\vspace{-4ex}
		\caption{Overview of our contrastive feature alignment (CFA) method. We first generate mixed clutter variants as references (Sec. \ref{cluttervarianst}), then the original images and their clutter variants are both inputs to the model. In the top inference branch, the channel-weighted MSE loss (Sec. \ref{cwmmse}) is presented to better measure/highlight the feature deviation resulting from clutter (\textit{contrast}) and extract similar deep features  (\textit{alignment}) from varying clutters. All three losses are jointly optimized to instruct the model in learning invariant representation of the target in various background clutters.}
		\label{frame}
		\vspace{-3ex}
	\end{figure*}

	In this letter, an invariant representation learning framework is proposed to obtain clutter robust SAR ATR models, namely, the contrastive feature alignment (CFA) method. The idea behind is to learn invariant target representation by contrasting and aligning the high-level features given rise by the original-reference sample pairs that carry different clutters. We propose a novel and effective mixed data augmentation strategy that makes use of both blank and noise backgrounds (clutter variants) as references to attract the deviation in feature space. And an additional inference branch equipped with the channel-weighted mean square error (CWMSE) loss function is proposed to better instruct the model in contrasting and aligning the deep feature for learning invariant target representation. Our method does not necessitate careful data processing in the test phase and can serve as an add-on to any well-trained models.
	The framework of the proposed CFA method is shown in Fig. \ref{frame} and our main contributions are summarized as follows.
	\begin{enumerate}
		\item We propose to fill the gap of SAR ATR between fixed background model training and varying background application in view of representation learning. By guiding the model to extract similar deep features from clutter variants, robust recognition of the target in unfamiliar clutters is achieved.
		
		\item The proposed CFA method customizes a clutter variants generation strategy that combines the blank and noise clutters as references to trigger deviation in feature space, and further devises an inference branch with CWMSE loss function to better contrast and align the features for learning the invariant target representation.
		
		\item Extensive experimental evaluations on six DNN models and the MSTAR dataset demonstrate the effectiveness and generalizability of the proposed method.
	\end{enumerate}

\section{Methodology}
\subsection{Motivation}\label{sec2}
This letter is established on the experimental observations that the clutter of SAR target chips roughly accounts for a third of the confidence score in DNN inference. We further found the model accuracy is severely corrupted when the target is surrounded by unfamiliar clutters, see Tables \ref{shapcomparason} and \ref{Noisecomparason}. Such a counter-intuitive distraction/overfitting and the resulting performance corruption raise concerns about their reliability in missions. 

One of the naive solutions for alleviating the overfitting on clutters is data augmentation, feeding models with the target images sensed in various surroundings. Whereas, the data collection is restricted by high cost or limited observing opportunities. Another is to train the model with target segments and thus neglect the clutters \cite{segclutter}. As mentioned before, it depends on the precision of segmentation techniques, which relates to image quality. 

The motivation behind this letter is to instruct DNNs in learning invariant representation of the target itself and performing robust decisions amongst varying background clutters without collecting massive training data or delicately processing the test data. Inspired by feature reconstruction \cite{yang2022self} and contrastive learning \cite{chen2020simple}, the CFA method is proposed  to achieve our goal. The proposed method is outlined in Fig. \ref{frame} and is elaborated on in the following.

\subsection{Overview}
Consider a SAR target classifier $F$ with parameters $\bm{\Theta}$ that $F(\cdot|\bm{\Theta}): \bm{x} \rightarrow y$ where $\bm{x}\in\mathbb{R}^{h\times w}$ is the SAR target image and $y\in \mathbb{R}^{1\times C}$ is the label over $C$ classes. Given an image $\bm{x}^{*}$ where the target $\bm{x}$ surrounded by another background clutter, we aim at obtaining the same correct prediction, that is, $\operatorname{argmax}F(\bm{x}^{*}|\bm{\Theta}) = \operatorname{argmax}F(\bm{x}|\bm{\Theta})$.

Our main idea is to optimize the parameters $\bm{\Theta}$ to enable the model capable of learning invariant representation by contrasting and aligning the deep features $\{f(\bm{x})$, $f(\tilde{\bm{x}})\}$ activated by the original-reference sample pairs $\{\bm{x}, \tilde{\bm{x}}\}$ that are different in terms of background clutters.
As shown in Fig. \ref{frame}, we first generate \textit{clutter variants} as references by retaining the target-related information (\textit{i.e.}, the target and shadow)
 and replacing the clutter with blank or random noise. Then, both the original images and corresponding clutter variants are fed into the backbone model to trigger deep features. Note that, in this stage, the deviation between original and reference features is caused by the variation of clutters. Assuming there exist both target-sensitive and clutter-sensitive features, we devise a CWMSE loss function to measure the channel-wise significance of the feature deviation (\textit{contrast}) and further decrease the deviation (\textit{align}) to obtain invariant target-related representation. Cooperating with the cross-entropy (CE) loss for classification, the model is trained to make correct decisions and extract similar deep features from the target for both the original images and clutter variants. And the learned invariant representation can generalize from the collection-free clutter variants to unknown measured clutters in mission.

\subsection{Clutter Variants Generation}\label{cluttervarianst}
To ensure that the learned feature representation can generalize well, the clutter variants need to be delicately designed. Generally, referring to the target with blank background is easier to find the robust target features than finding shared features in different clutters. Whereas, on the other hand, the noise clutter variants are also valuable as data augmentation for a model to fit the varying environment. Thus, we propose to generate mixed clutter variants including both blank and noise clutter variants as references.

Considering the general mini-batch training process, given a batch of training data points $\bm{X}_{B}=\{\bm{x}_{1}, \bm{x}_{2}, \cdots, \bm{x}_{b}\}$ with labels $\bm{y}_{B}=\{y_{1},y_{2},\cdots,y_{b}\}$, the corresponding clutter variants $\tilde{\bm{X}}_{B}$ are obtained by replacing the original clutters with log-normal distributed noise \cite{george1968detection} or blank clutter with probability $p$ as
\begin{equation}
\begin{aligned}
\tilde{\bm{X}}_{B} &= \{\tilde{\bm{x}}_{1},\tilde{\bm{x}}_{2}, \cdots, \tilde{\bm{x}}_{b}\}, \\
\tilde{\bm{x}}_{i} &= \begin{cases} \bm{x}_{i}\!\odot\!(\mathbf{m}_{i,t}\!+\!\mathbf{m}_{i,s}) \!+\! \bm{n} \! \odot \! \mathbf{m}_{i,c}, \text{with probability} \;  p \\ 
\bm{x}_{i}\!\odot\!(\mathbf{m}_{i,t}\!+\!\mathbf{m}_{i,s}),   \text{with probability}  \;  1-p 
\end{cases}\\
\bm{n} &\sim \mathcal{N}_{\text{log}}(n_{m}, n_{\sigma}),
\end{aligned}
\end{equation}
where $\mathbf{m}_{t/s/c}$ are respectively the segmentation mask for the target, shadow, and clutter region, $n_{m}$ and $n_{\sigma}$ are respectively the mean value and standard deviation of the log-normal distribution $\mathcal{N}_{\text{log}}$. In experiments, we set $n_{m} \sim U(0.1, 0.2)$ and $n_{\sigma} \sim U(0.1, 0.3)$. The randomness is designed to trigger diverse reference features. In this letter, only a single reference image is sampled once a time and different samples across training epochs can bring enough diversity.

\begin{table*}[tbp]
	\centering	
	\caption{Average Shapley values (\%) of each part of the image over the test set. We first segment obtain the target, shadow, and clutter segments $\{\bm{x}_{t},\bm{x}_{s},\bm{x}_{c}\}$ with the annotations provided by SARBake dataset \cite{malmgren2015convolutional}. Then, we calculate the Shapley value \cite{shap} of each part's contribution for the normalized confidence score}
	\label{shapcomparason}
	\vspace{-2ex}
	\begin{tabular}{ccccccccccccc}
		\toprule[1.5pt]
		\multirow{3}{*}{Model} & \multicolumn{2}{c}{AConvNet \cite{chen2016target} }  & \multicolumn{2}{c}{AM-CNN \cite{zhang2020convolutional}} & \multicolumn{2}{c}{MVGGNet \cite{zhang2020fec}} & \multicolumn{2}{c}{ResNet18 \cite{resnet2016kh}} & \multicolumn{2}{c}{ShuffleNetV2 \cite{Ma_2018_ECCV}} & \multicolumn{2}{c}{MobileNetV2 \cite{mobilev2}} \\ 	\cmidrule(r){2-3}  \cmidrule(r){4-5}  \cmidrule(r){6-7}  \cmidrule(r){8-9}  \cmidrule(r){10-11}  \cmidrule(r){12-13} 
		& Baseline & \textbf{Ours} & Baseline & \textbf{Ours}& Baseline & \textbf{Ours} & Baseline & \textbf{Ours}& Baseline & \textbf{Ours} & Baseline & \textbf{Ours} \\ \cmidrule(r){2-13} 
		Accuracy & 97.9 & 98.0  & 97.7 & 97.6 & 97.0 &  96.5 & 98.4 & 97.1 & 96.1 & 95.5 & 98.0 & 97.8 \\ \midrule
		Target$\uparrow$ & 53.8 & 68.3  & 42.3 & 73.4 & 39.9 &  59.3 & 39.0 & 68.8 & 32.1 & 56.3 & 36.1 & 56.5 \\ \cmidrule(r){2-13} 
		Shadow$\uparrow$ & 20.9 & 23.1  & 27.2 & 17.3 & 26.4 & 31.9 & 29.1 & 24.8 & 31.1 & 32.1 & 28.6 & 32.2  \\ \cmidrule(r){2-13} 
		Clutter$\downarrow$  & 24.0 & 6.0 & 27.8 & 6.5 & 32.8 & 6.7 & 30.4 & 4.7 & 34.8 & 8.9 & 44.0 & 8.6   \\ 
		\bottomrule[1.5pt]
	\end{tabular}
\vspace{-2ex}
\end{table*}%

\subsection{Channel-Weighted MSE Loss}\label{cwmmse}
During the training process, the original images and corresponding clutter variants both flow through the layers, we hook the original features $f(\bm{X}_{B})$ and reference features $f(\tilde{\bm{X}}_{B})$ which share a shape of $b\times c\times h_{f} \times w_{f}$. Recall that there exist target-sensitive and clutter-sensitive features, that is, when the clutter changes, the target-sensitive features are stable while the clutter-sensitive features vary a lot. Therefore, we weight the MSE loss in channel dimension based on the distance between original and reference features in each channel:
\begin{equation}
\begin{aligned}
\mathcal{L}_{\text{CWMSE}}&(f(\bm{X}_{B}), f(\tilde{\bm{X}}_{B})) \\
&= \frac{1}{b c h_{f} w_{f}} \sum_{i}^{b}\sum_{j}^{c}w_{i,j}\sum_{k}^{h_{f}}\sum_{l}^{w_{f}}(f(\bm{x}_{i})-f(\tilde{\bm{x}}_{i}))^{2},
\end{aligned}
\end{equation} 
where $w_{i,j}$ is the weight of the $i$-th sample in $j$-th channel and is calculated by
\begin{equation}
w_{i,j}=\frac{c\sum_{k}^{h_{f}}\sum_{l}^{w_{f}}(f(\bm{x}_{i})-f(\tilde{\bm{x}}_{i}))^{2}}{ h_{f} w_{f}\sum_{j}^{c}\sum_{k}^{h_{f}}\sum_{l}^{w_{f}}(f(\bm{x}_{i})-f(\tilde{\bm{x}}_{i}))^{2}}.
\end{equation} 
The above design satisfies $\frac{1}{c}\sum_{j}^{c}w_{i,j}=1$, and the most similar feature channels get small weights while the unlike channels gain larger ones, enabling the model to focus more on the clutter-sensitive features.

\subsection{Training Strategy}
The model is jointly trained by CE and CWMSE losses, and the total loss function is 
\begin{equation}
\mathcal{L}_{\text{total}} \!\! = \! \! \underbrace{\frac{1}{2}\!(\mathcal{L}_{\text{CE}}(F(\bm{X}),\!\bm{y})\!\! +\!\! \mathcal{L}_{\text{CE}}(F(\tilde{\bm{X}}),\!\bm{y}))}_{\text{Clasification}} \! + \! \underbrace{\lambda\mathcal{L}_{\text{CWMSE}}(f(\bm{X}),\! f(\tilde{\bm{X}}))}_{\text{Contrastive feature alignment}},
\end{equation} 
where $\lambda$ is the balance factor. 
Mathematically, $\mathcal{L}_{\text{CWMSE}}$ limits the solution space of the two joint CE losses but helps to learn the invariant representation. Deep features at the last convolution layer are selected to optimize $\mathcal{L}_{\text{CWMSE}}$ due to their high-level representation ability.

To better trigger and further distinguish the clutter-sensitive features, we use well-trained classifiers as the backbone. That is, the initial parameters perform well on test images but overfit the target-clutter correlation. We train the model with a relatively large initial learning rate, \textit{e.g.}, 0.5, and sequentially fine-tune it to obtain stable parameters. 

\begin{figure*}[tbp]
	\centering
	\includegraphics[height=0.225\linewidth]{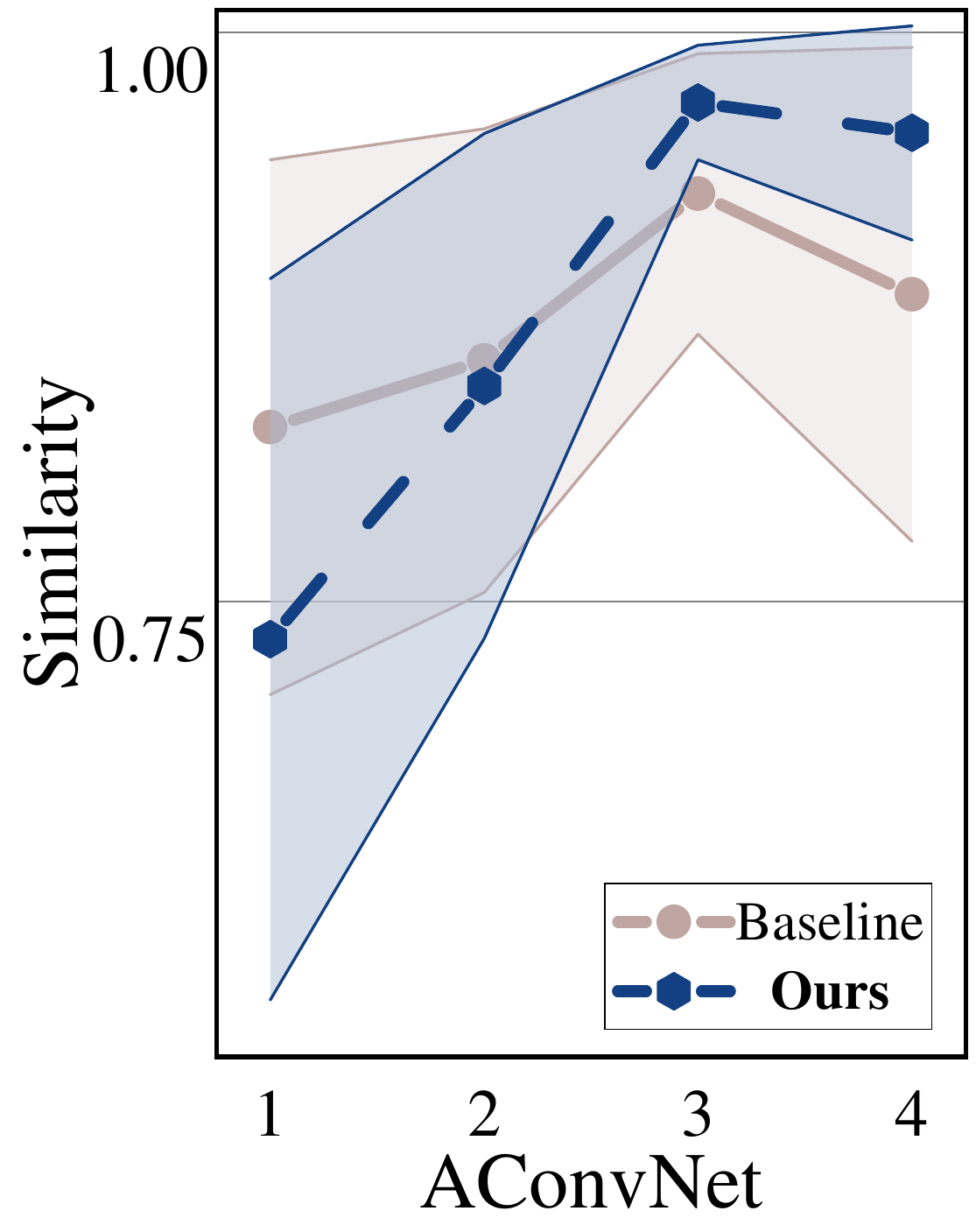}\!\!
	\includegraphics[height=0.225\linewidth]{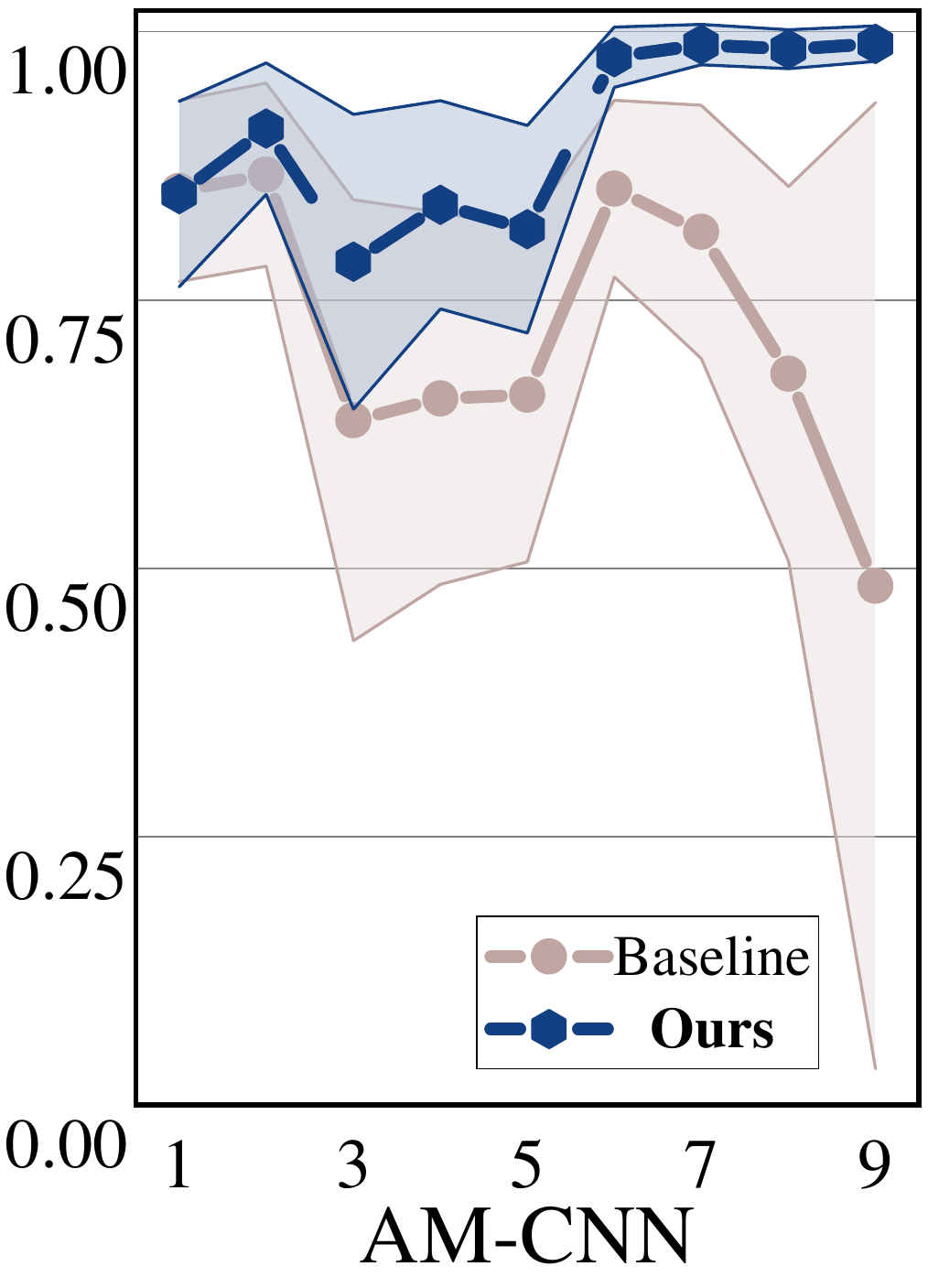}\!\!
	\includegraphics[height=0.225\linewidth]{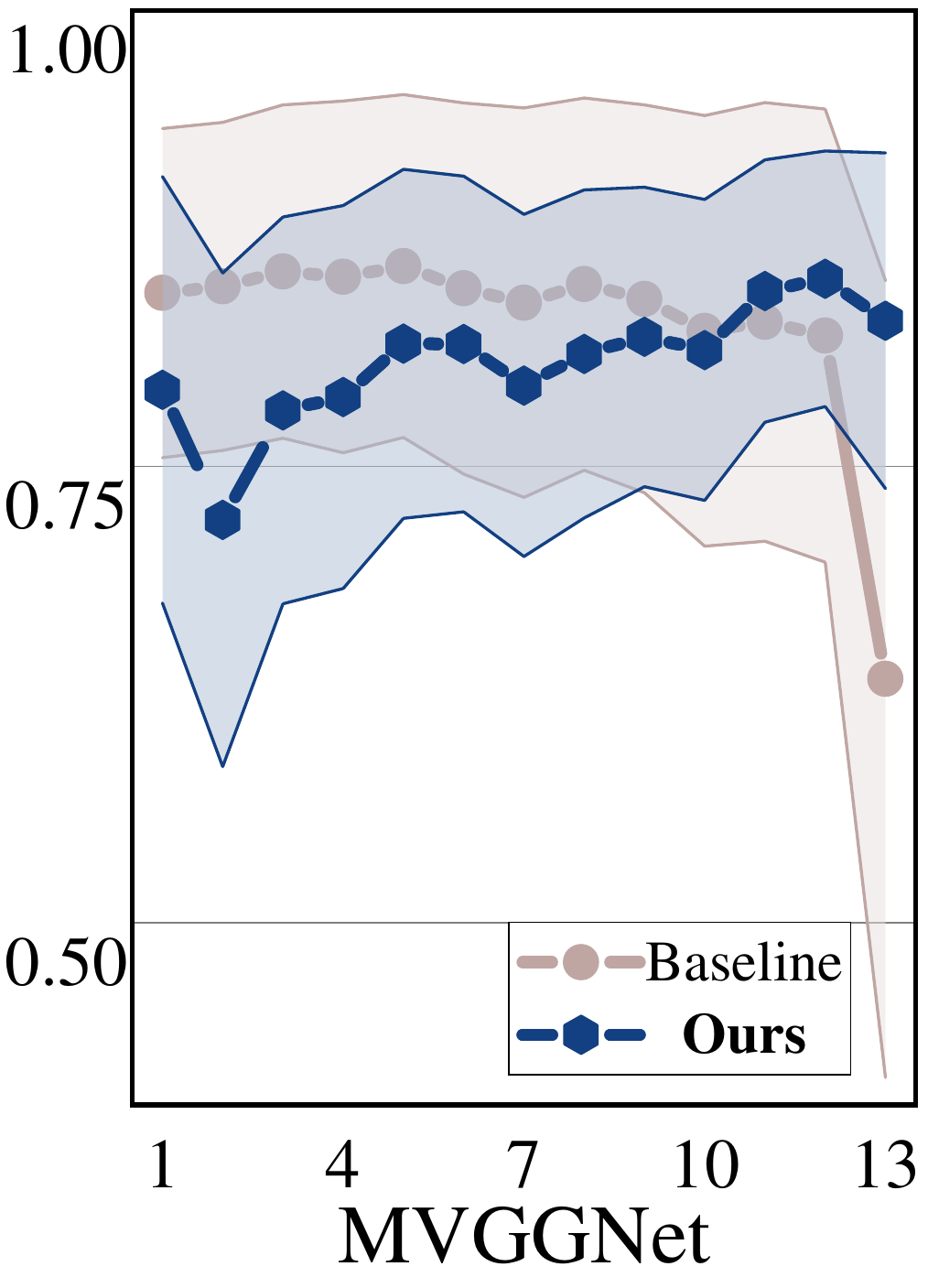}\!\!
	\includegraphics[height=0.225\linewidth]{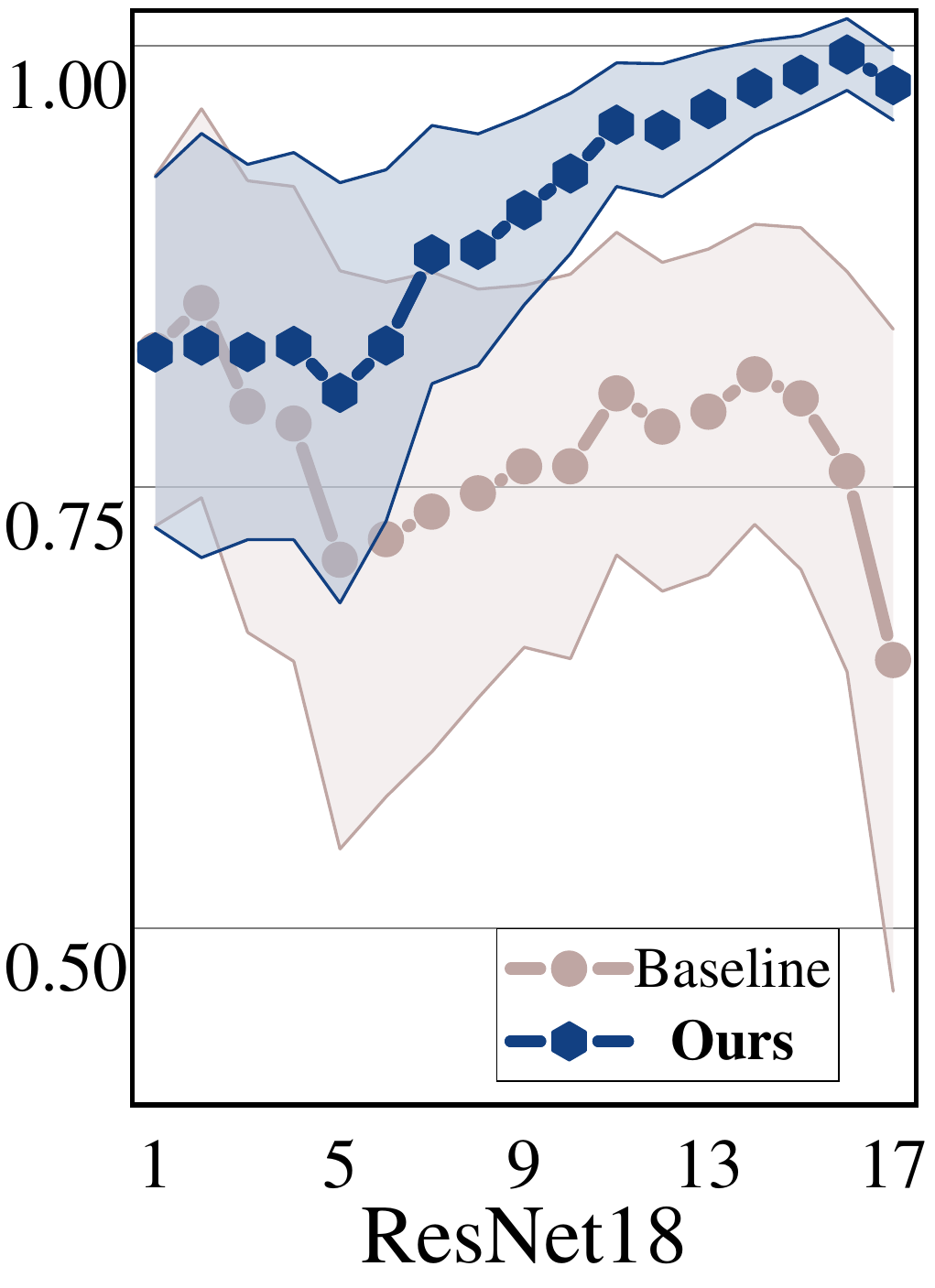}\!\!
	\includegraphics[height=0.225\linewidth]{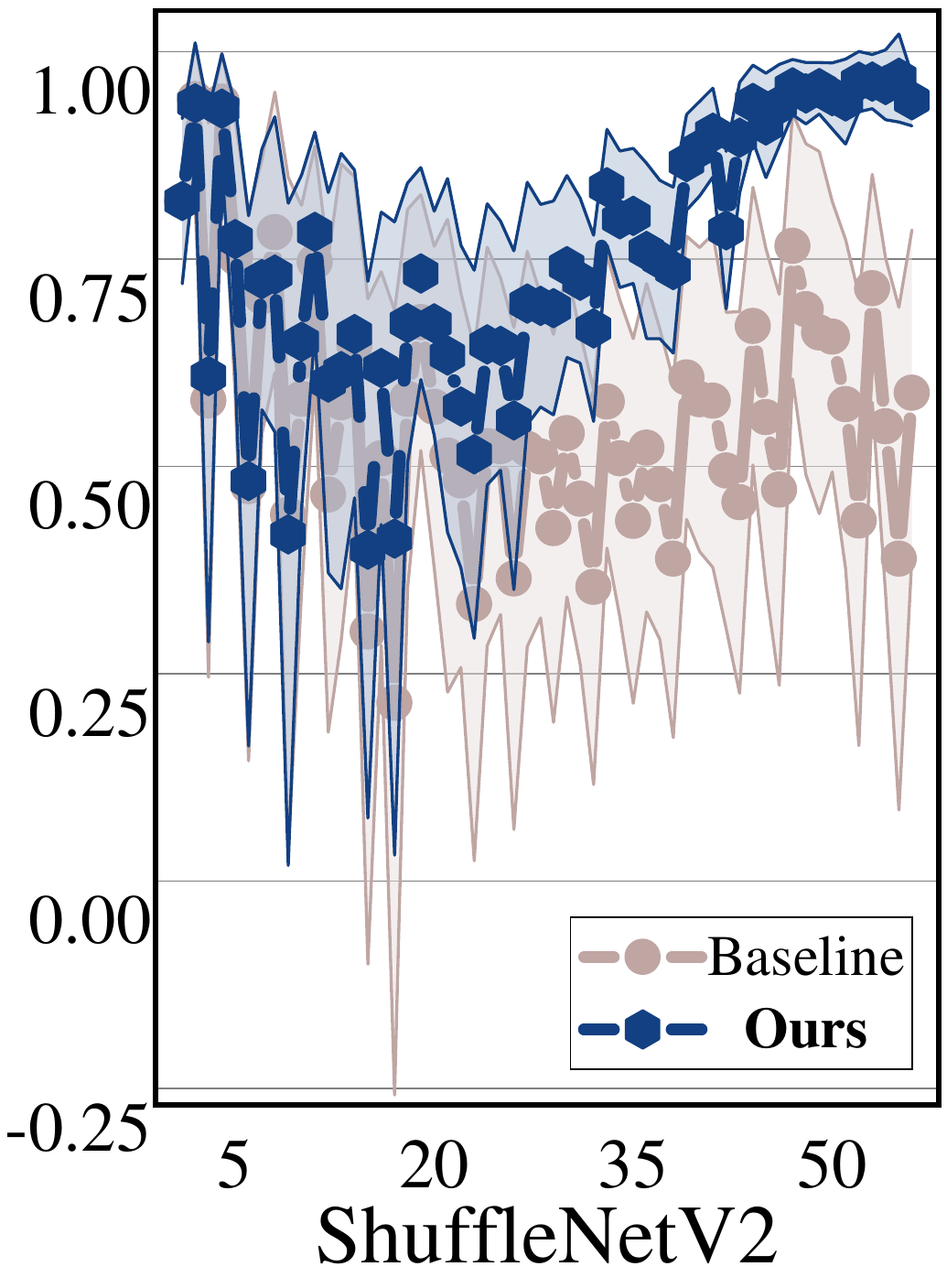}\!\!
	\includegraphics[height=0.225\linewidth]{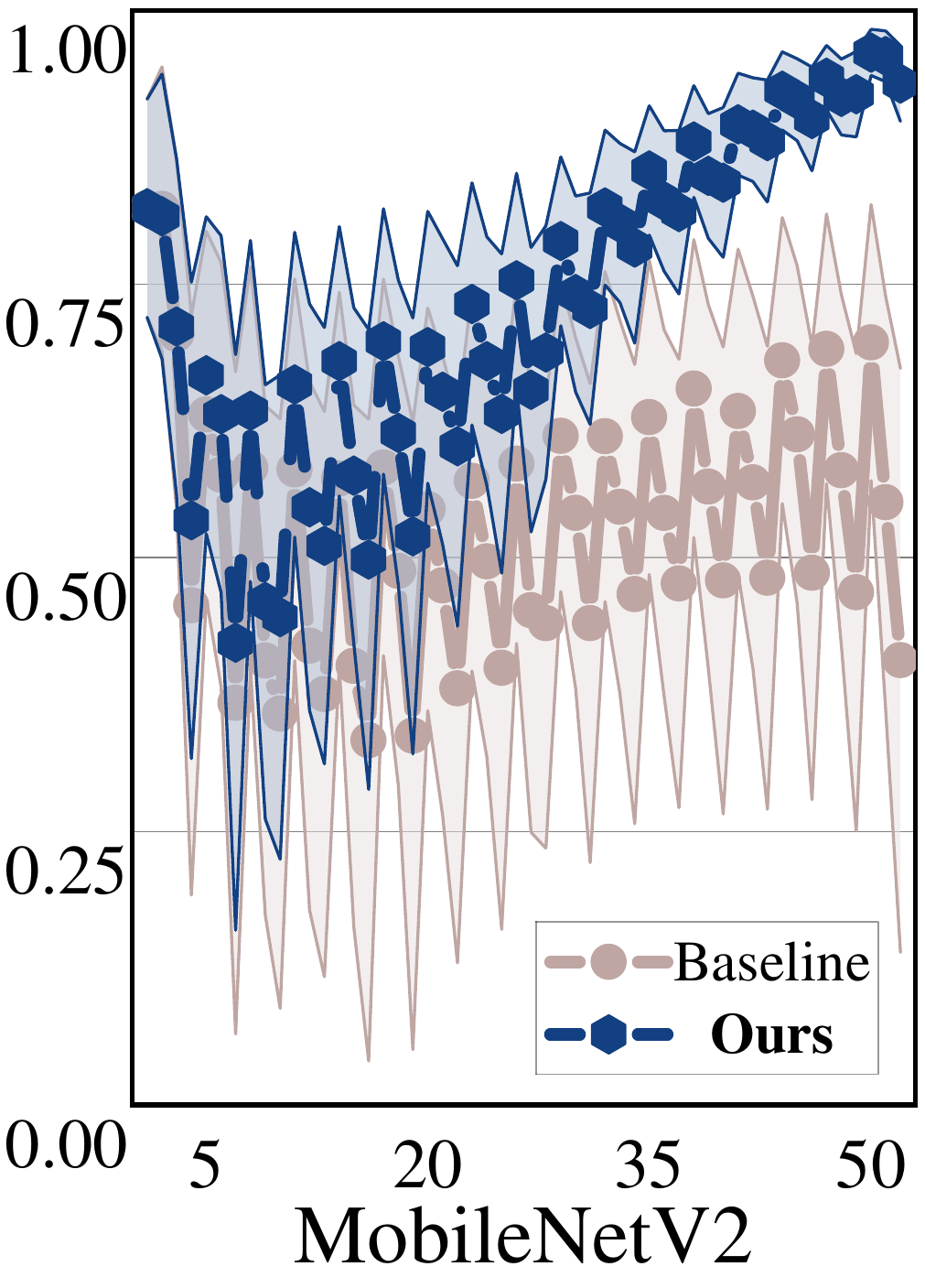}
		\vspace{-2ex}
	\caption{Layer-wise cosine similarity between interior features at convolution layers respectively given rise by the test images and its real-clutter variants. The results were averaged over the test set on 10 runs.}
	\label{cossim}
	\vspace{-4ex}
\end{figure*}

\section{Experiments}\label{experiments}
\subsection{Setup}\label{setup}
\subsubsection{Dataset and Models}
The experimental evaluations were conducted upon the MSTAR dataset, which was imaged at X-band with 1-ft resolution and made publicly accessible by the moving and stationary target acquisition and recognition program. Ten categories of targets were included to train AConvNet \cite{chen2016target}, AM-CNN \cite{zhang2020convolutional}, MVGGNet \cite{zhang2020fec}, ResNet18 \cite{resnet2016kh}, ShuffleNetV2 \cite{Ma_2018_ECCV}, and MobileNetV2 \cite{mobilev2} for attaining representative results. Note that, the first three models were specially designed for SAR ATR task and the last three are general backbones for optical image classification. Specifically, the central patches with a pixel size of $88\times 88$ were for training ($17^{\circ}$ depression angle) and testing ($15^{\circ}$ depression angle) the models, and the capacity of the training set was 2747, while 2425 for the test set. To better investigate the effectiveness of our method, we trained the models without other data augmentation, such as rotation, additive noise, \textit{etc.} In experiments we set $\lambda=1$ and $p=0.8$ based on the results reported in Sec. \ref{ablationstudy}.

\subsubsection{Baseline}
As the CE loss function has been proved efficient and effective in both optical and SAR image target recognition tasks, and the proposed method serves as an add-on for additional performance in alleviating the clutter-target correlation influence. 
We investigated the effectiveness of the CFA method by comparing the models respectively trained by the CE loss equipped with and without our proposal, especially in terms of target focus and robustness against clutter replacement.

\subsubsection{Measurements}
We measured the effectiveness of the models in robust recognition by the decrease of accuracy (\%). There were mainly three test conditions: 1) test images with replaced clutters randomly sampled from the large scene image provided by the MSTAR dataset, 2) test images with higher SCR (+3dB), and 3) test images with lower SCR (-3dB). The measured clutters were collected by the same sensor as the one collected the target chips, and the SCR fluctuation was implemented by tuning the clutter magnitude. Corresponding test images are illustrated in Fig. \ref{gbp}.

\subsection{Results} \label{results}
Initially, the effectiveness of our method was examined. Table \ref{shapcomparason} lists the models' accuracy and Shapley values for each part of the SAR target images. From the data, one should notice there existed a slight decrease in the accuracy, which is expectable as the models intentionally neglect the discriminative information carried by clutters. And also due to that, our models took advantage of the target focusing which is shown by the significant increase in the Shapley values of the target and shadow parts to perform correct decisions. In particular, for all baseline models, the Shapley values of the clutter are above 20\%, especially the values of ShuffleNetV2 and MobileNetV2 are greater than those of the target. With training by our method, the Shapley values of the target increased by 23.2\%, the clutter decreased by 25.4\% on average, and all the clutter values are below 10\%.

Further, we calculated the average cosine similarity of interior features between the test images and their real-clutter variants over the test set, with results shown in Fig. \ref{cossim}. From these curves, one could observe that the initial similarities for all models were about 0.8, along the layers our models unremittingly extracted the correct robust features while the baseline models got lost in the unknown clutters layer by layer.

Then, the robustness of our method was verified with details elaborated in Table \ref{Noisecomparason}, from which the following observations can be concluded. First, besides being robust to the blank and noise clutters that we had fed into training, the models can also make consistent decisions in facing the real-clutter variants, demonstrating the satisfactory generalizability of the proposed method. Second, according to \cite{weijie}, the baseline models partially overfit the fixed and comparable SCRs across classes, and models trained by our scheme gave evidence of the positive utility in being robust to different SCR fluctuations. 
\begin{figure*}[htbp]
	\centering
	\includegraphics[width=1\linewidth]{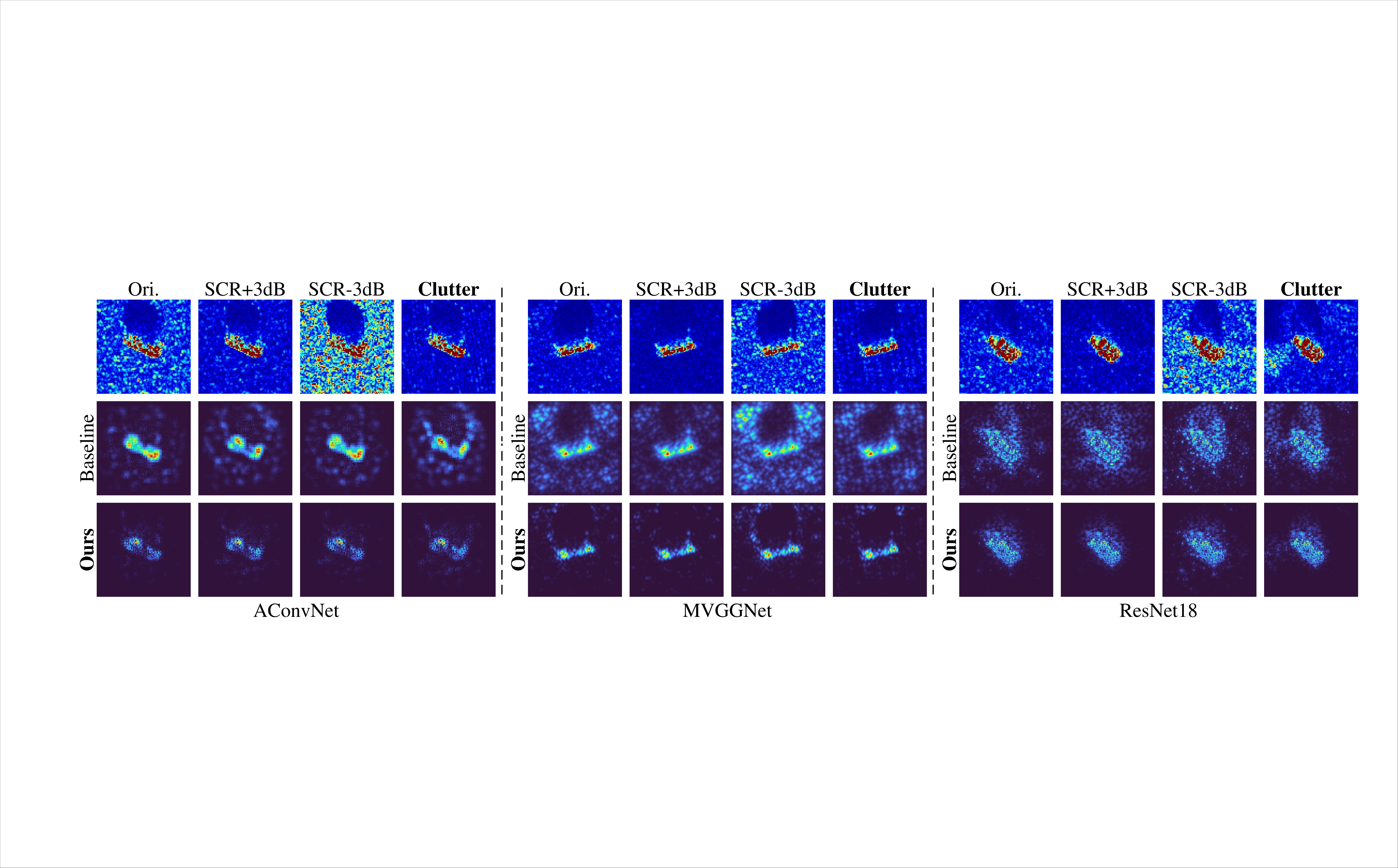}
	\vspace{-5ex}
	\caption{Guided-Backpropagation \cite{springenberg2014striving} results for AConvNet, MVGGNet, and ResNet18, respectively. In each subgroup, the original image, real-clutter variant, and two SCR fluctuation variants are arranged in the first row (from left to right), the corresponding attention maps extracted from the baseline and our models are depicted below.}
	\label{gbp}
	\vspace{-3ex}
\end{figure*}
\begin{table}
	\centering
	\caption{Decrease of accuracy (\%) under interference: 1) clutter replaced by null matrix (Blank), log-normal distributed noise with mean value of 0.15 and standard deviation of 0.2 (Noise), and real clutters (\textbf{Clutter}); 2) fluctuation of the signal-to-clutter ratio (SCR), implemented by tuning the clutter magnitude. All the results of experiments with randomness were averaged on 10 runs}
	\label{Noisecomparason}
	\vspace{-2ex}
	\begin{tabular}{ccccccc}
		\toprule[1.5pt]
		& & \multicolumn{3}{c}{Replacements} &  \multicolumn{2}{c}{SCR fluctuation}   \\ 	\cmidrule(r){3-5}  \cmidrule(r){6-7} 
		Model  & Loss & Blank & Noise & \textbf{Clutter} & 3dB & -3dB  \\ \midrule	
		\multirow{2}{*}{AConvNet} & Baseline & 55.5  & 27.7 & 47.6 & 32.1  & 50.8      \\
		& \textbf{Ours} & 0.8 & 0.8 & 0.9 & 0.3 & 4.8 \\ \cmidrule(r){2-7} 
		\multirow{2}{*}{AM-CNN} & Baseline & 62.4  & 28.0 & 48.9 & 23.8 & 34.6  \\
		& \textbf{Ours} & 0.4 & 0.3 & 0.5 & 0.3 & 3.4  \\ \cmidrule(r){2-7} 
		\multirow{2}{*}{MVGGNet} & Baseline & 78.2  & 53.4 & 69.7 & 63.1 & 72.7   \\
		& \textbf{Ours} & -0.7 & -0.1 & -0.1 & -0.5 & 2.1   \\  \cmidrule(r){2-7} 
		\multirow{2}{*}{ResNet18} & Baseline & 73.4  & 68.0 & 56.7 & 34.8  & 51.0     \\
		& \textbf{Ours} & 3.1 & 0.8 & 1.1 & 0.9 & 17.5  \\ \cmidrule(r){2-7} 
		\multirow{2}{*}{ShuffleNetV2} & Baseline & 85.9  & 41.5 & 72.7 & 55.3 &  68.4   \\
		& \textbf{Ours} & 6.3 & 2.3 & 4.8 & 3.2 & 19.8   \\ \cmidrule(r){2-7} 
		\multirow{2}{*}{MobileNetV2} & Baseline & 83.8  & 58.8 & 64.8 & 36.4 & 54.4  \\
		& \textbf{Ours} & 3.4 & 1.4 & 3.7 & 1.1 & 21.5   \\ 
		\bottomrule[1.5pt]
	\end{tabular}
	\vspace{-5ex}
\end{table}

Fig. \ref{gbp} illustrates the model attention by utilizing the Guided-Backpropagation \cite{springenberg2014striving} in which the brighter pixels were more discriminative to the models. From these maps, one can aware that the baseline models captured the information contained in the clutter, and were interfered by SCR fluctuations and unfamiliar clutters, whereas our models were indifferent to the varying clutters and consistently focused on the target. 

\subsection{Ablation Study and Comparative Experiments}\label{ablationstudy}
\begin{figure}[tbp]
	\centering
	\includegraphics[height=0.45\linewidth]{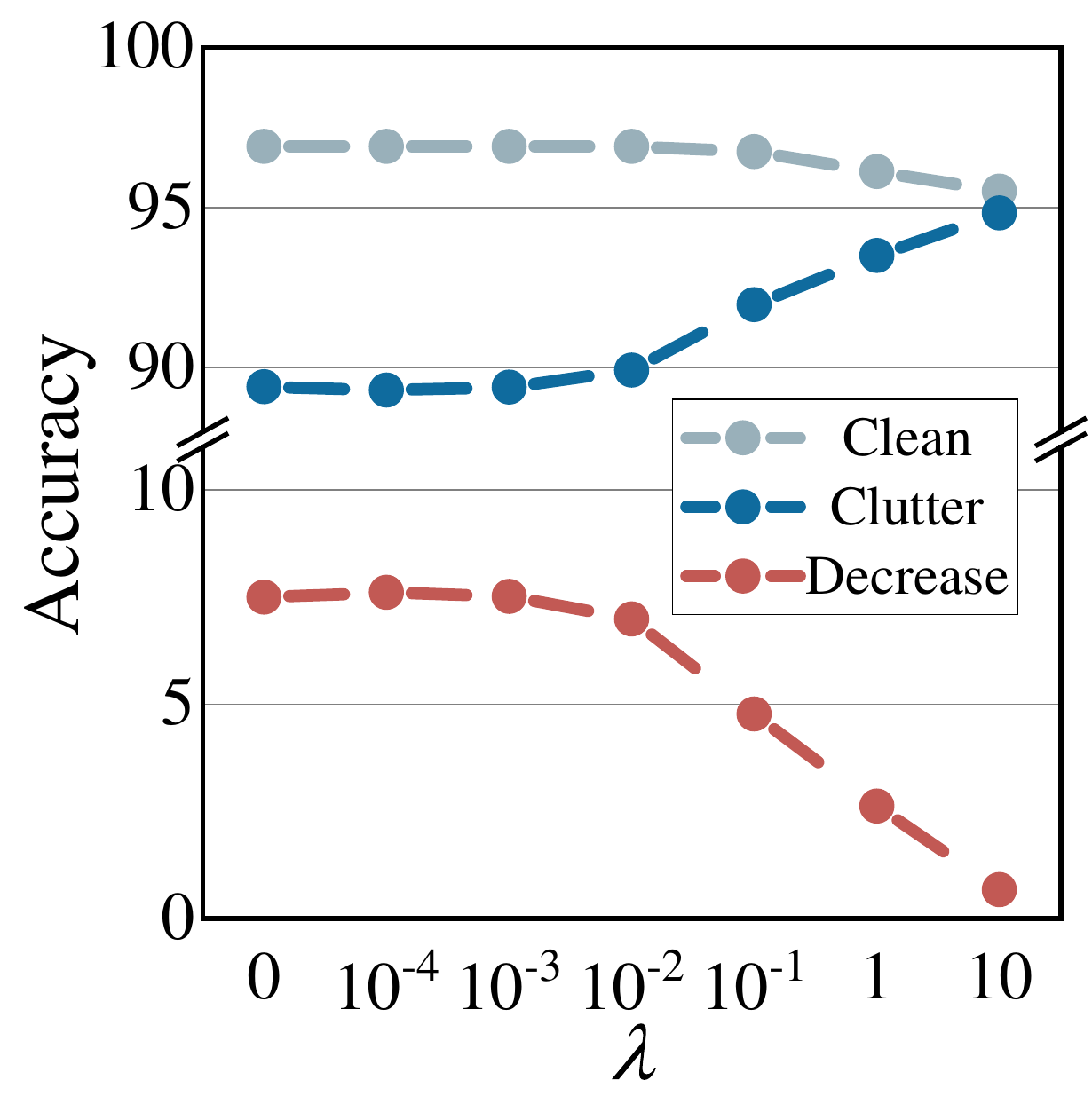}
	\includegraphics[height=0.45\linewidth]{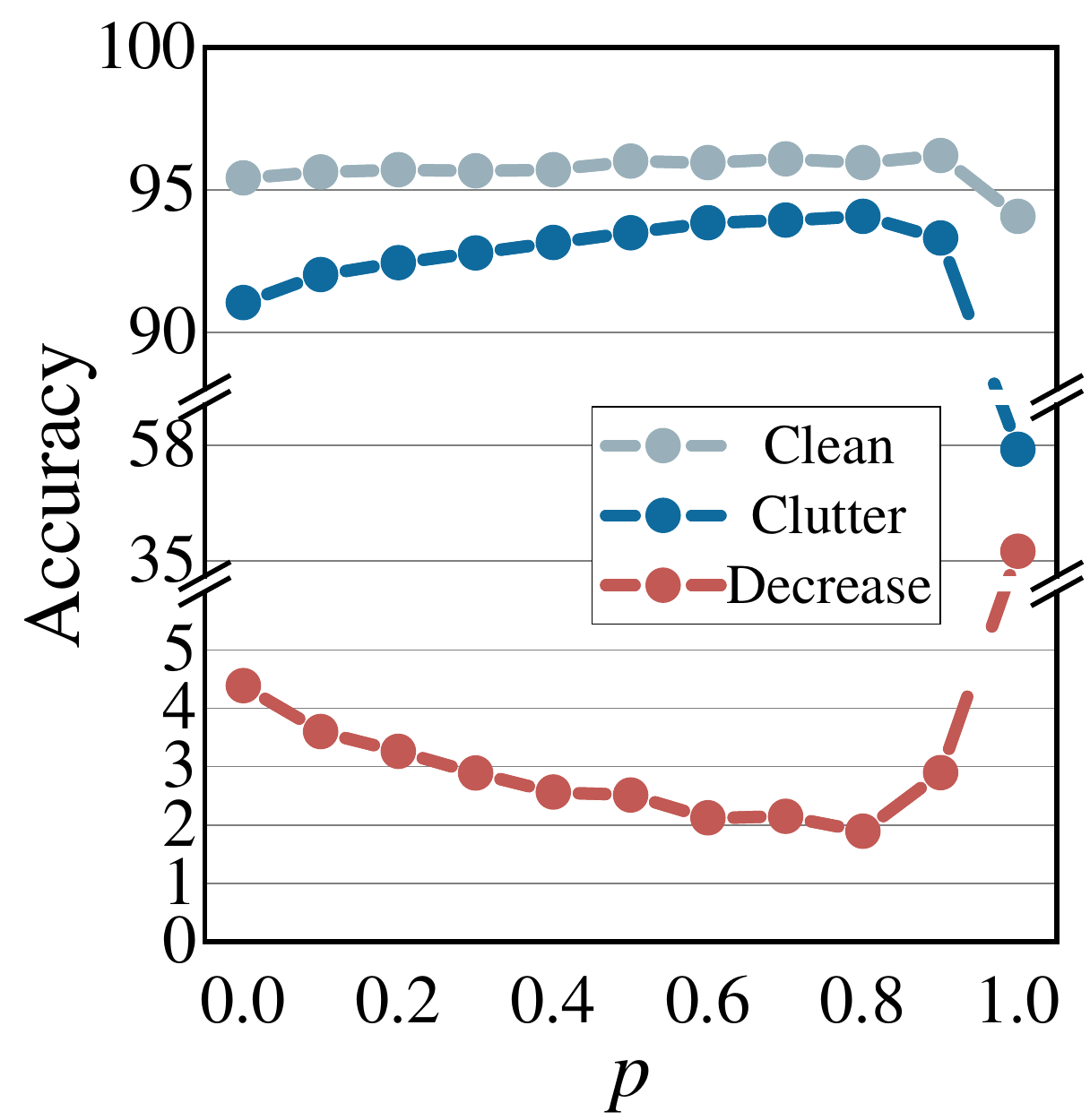}
	\vspace{-2ex}
	\caption{Accuracy (\%) on the original test images, real-clutter variants, and the accuracy decrease as functions of $\lambda$ and $p$. The ResNet18 was trained with the same epochs and learning rates under fixed random seed. We studied the effect of $\lambda$ with $p=0.5$, and $p$ with $\lambda=1$.}
	\label{ablation}
	\vspace{-2ex}
\end{figure}
We report the results of the ablation study, see Fig. \ref{ablation} for details. Results showed the effectiveness of our contrastive feature alignment and the combined clutter variants generation strategies. In specific, the contrastive feature alignment strategy outperformed the data augmentation method ($\lambda=0$) and brought significant accuracy increment for real-clutter variants (over 4\%). Combining blank and noise clutter variants as references also showed over 2\% and 37\% increments compared with using a single reference. Moreover, $\mathcal{L}_{\text{CWMSE}}$ gained 0.12\% and 0.26\% accuracy increments on original test images and real-clutter variants respectively to the vanilla MSE loss when $\lambda=1$ and $p=0.8$.

We then compared our method with the target segments-based solution with results shown in Table \ref{comparativeexps}. As can be seen from the results, the method suggested in \cite{segclutter} (Competitor \#1) can totally neglect the influence of clutter due to it utilizing the aforehand segmentation as inappropriate prior for testing. However, we found it suffers from the segmentation results in the application, which is affected by the image quality or SCR fluctuations (Competitor \#2), proving the advantage of our proposal.

\begin{table}
	\centering
	\caption{Decrease of accuracy (\%) in comparative experiment. Competitor \#1 was trained with SARBake annotations and Competitor \#2 was trained with target segments obtained by the method in \cite{sun2007adaptive}}
	\label{comparativeexps}
	\vspace{-2ex}
	\begin{tabular}{ccccc}
		\toprule[1.5pt]
		ResNet18 & Accuracy & \textbf{Clutter} &SCR+3dB & SCR-3dB \\ \midrule
		Competitor \#1  & 97.3 & 0 & 0 & 0  \\ \midrule
		Competitor \#2   & 95.2 & 40.9 & 16.1 & 44.8 \\ \midrule
		\textbf{Ours}  & 97.1 & 1.1 & 0.9 & 17.5 \\
		\bottomrule[1.5pt]
	\end{tabular}
	\vspace{-3ex}
\end{table}

\section{Conclusion and Discussion}\label{conclusion}

This letter has proposed a CFA method to learn invariant presentation for clutter robust SAR ATR. Both the mixed data augmentation and channel-weighted distance measurement were devised to better train models in learning the target representation and being robust to clutter and SCR variations. Our method can also serve as an add-on for other application scenarios such as recognition for the interfered target. 
Extensive experiments on the MSTAR dataset demonstrated that our proposal is effective for generalizable and robust SAR ATR applications.

It also raises concerns about whether highly target-focused models are more vulnerable to target-specific interference. For instance, current studies demonstrated that the DNN-based SAR ATR models are surprisingly beaten by the imperceptible yet malicious adversarial perturbations \cite{SVA,smgaa}. We include the adversarial risks of our method in upcoming work.
	
	\bibliographystyle{IEEEtran}    
	\bibliography{refediffasc}                   

\begin{thebibliography}{10}
\providecommand{\url}[1]{#1}
\csname url@samestyle\endcsname
\providecommand{\newblock}{\relax}
\providecommand{\bibinfo}[2]{#2}
\providecommand{\BIBentrySTDinterwordspacing}{\spaceskip=0pt\relax}
\providecommand{\BIBentryALTinterwordstretchfactor}{4}
\providecommand{\BIBentryALTinterwordspacing}{\spaceskip=\fontdimen2\font plus
\BIBentryALTinterwordstretchfactor\fontdimen3\font minus
  \fontdimen4\font\relax}
\providecommand{\BIBforeignlanguage}[2]{{%
\expandafter\ifx\csname l@#1\endcsname\relax
\typeout{** WARNING: IEEEtran.bst: No hyphenation pattern has been}%
\typeout{** loaded for the language `#1'. Using the pattern for}%
\typeout{** the default language instead.}%
\else
\language=\csname l@#1\endcsname
\fi
#2}}
\providecommand{\BIBdecl}{\relax}
\BIBdecl

\bibitem{kechagias2021automatic}
O.~Kechagias-Stamatis and N.~Aouf, ``{Automatic Target Recognition on Synthetic
  Aperture Radar Imagery: A survey},'' \emph{IEEE Aerospace and Electronic
  Systems Magazine}, vol.~36, no.~3, pp. 56--81, 2021.

\bibitem{schumacher2005non}
R.~Schumacher and J.~Schiller, ``Non-cooperative target identification of
  battlefield targets-classification results based on sar images,'' in
  \emph{IEEE International Radar Conference, 2005.}\hskip 1em plus 0.5em minus
  0.4em\relax IEEE, 2005, pp. 167--172.

\bibitem{schumacher2004atr}
R.~Schumacher and K.~Rosenbach, ``{ATR of Battlefield Targets by SAR
  Classification Results Using the Public MSTAR Dataset Compared with a Dataset
  by QinetiQ UK},'' in \emph{RTO SET Symposium on Target Identification and
  Recognition Using RF Systems}.\hskip 1em plus 0.5em minus 0.4em\relax
  Citeseer, 2004.

\bibitem{clutter}
F.~Zhou, L.~Wang, X.~Bai, and Y.~Hui, ``{SAR ATR of Ground Vehicles Based on
  LM-BN-CNN},'' \emph{IEEE Transactions on Geoscience and Remote Sensing},
  vol.~56, no.~12, pp. 7282--7293, 2018.

\bibitem{AnylisiTAES}
C.~Belloni, A.~Balleri, N.~Aouf, J.-M. Le~Caillec, and T.~Merlet,
  ``Explainability of deep sar atr through feature analysis,'' \emph{IEEE
  Transactions on Aerospace and Electronic Systems}, vol.~57, no.~1, pp.
  659--673, 2021.

\bibitem{weijie}
\BIBentryALTinterwordspacing
W.~Li, W.~Yang, L.~Liu, and Y.~Liu. Discovering and explaining the
  non-causality of deep learning in sar atr. [Online]. Available:
  \url{https://arxiv.org/abs/2304.00668}
\BIBentrySTDinterwordspacing

\bibitem{segclutter}
M.~Heiligers and A.~Huizing, ``On the importance of visual explanation and
  segmentation for sar atr using deep learning,'' in \emph{2018 IEEE Radar
  Conference (RadarConf18)}, 2018, pp. 0394--0399.

\bibitem{malmgren2015convolutional}
D.~Malmgren-Hansen and M.~Nobel-J, ``Convolutional neural networks for sar
  image segmentation,'' in \emph{Proceedings of the IEEE Symposium on Signal
  Processing and Information Technology}, 2015, pp. 231--236.

\bibitem{yang2022self}
Z.~Yang, W.~Dong, X.~Li, J.~Wu, L.~Li, and G.~Shi, ``Self-feature distillation
  with uncertainty modeling for degraded image recognition,'' in \emph{Computer
  Vision--ECCV 2022: 17th European Conference, Tel Aviv, Israel, October
  23--27, 2022, Proceedings, Part XXIV}, 2022, pp. 552--569.

\bibitem{chen2020simple}
T.~Chen, S.~Kornblith, M.~Norouzi, and G.~Hinton, ``A simple framework for
  contrastive learning of visual representations,'' in \emph{International
  conference on machine learning}.\hskip 1em plus 0.5em minus 0.4em\relax PMLR,
  2020, pp. 1597--1607.

\bibitem{george1968detection}
S.~F. George, ``The detection of nonfluctuating targets in log-normal
  clutter,'' NAVAL RESEARCH LAB WASHINGTON DC, Tech. Rep., 1968.

\bibitem{shap}
L.~S. Shapley, ``A value for n-person games,'' \emph{Contributions to the
  Theory of Games II}, vol.~28, pp. 307--317, 1953.

\bibitem{chen2016target}
S.~Chen, H.~Wang, F.~Xu, and Y.-Q. Jin, ``Target classification using the deep
  convolutional networks for sar images,'' \emph{IEEE Transactions on
  Geoscience and Remote Sensing}, vol.~54, no.~8, pp. 4806--4817, 2016.

\bibitem{zhang2020convolutional}
M.~Zhang, J.~An, L.~D. Yang, L.~Wu, X.~Q. Lu \emph{et~al.}, ``Convolutional
  neural network with attention mechanism for sar automatic target
  recognition,'' \emph{IEEE geoscience and remote sensing letters}, vol.~19,
  pp. 1--5, 2020.

\bibitem{zhang2020fec}
J.~Zhang, M.~Xing, and Y.~Xie, ``Fec: A feature fusion framework for sar target
  recognition based on electromagnetic scattering features and deep cnn
  features,'' \emph{IEEE Transactions on Geoscience and Remote Sensing},
  vol.~59, no.~3, pp. 2174--2187, 2020.

\bibitem{resnet2016kh}
K.~He, X.~Zhang, S.~Ren, and J.~Sun, ``Deep residual learning for image
  recognition,'' in \emph{Proceedings of the IEEE Conference on Computer Vision
  and Pattern Recognition}, 2016, pp. 770--778.

\bibitem{Ma_2018_ECCV}
N.~Ma, X.~Zhang, H.-T. Zheng, and J.~Sun, ``Shufflenet v2: Practical guidelines
  for efficient cnn architecture design,'' in \emph{Proceedings of the European
  Conference on Computer Vision}, 2018.

\bibitem{mobilev2}
M.~Sandler, A.~Howard, M.~Zhu, A.~Zhmoginov, and L.-C. Chen, ``Mobilenetv2:
  Inverted residuals and linear bottlenecks,'' in \emph{Proceedings of the IEEE
  Conference on Computer Vision and Pattern Recognition}, 2018, pp. 4510--4520.

\bibitem{springenberg2014striving}
J.~T. Springenberg, A.~Dosovitskiy, T.~Brox, and M.~Riedmiller, ``Striving for
  simplicity: The all convolutional net,'' in \emph{Proceedings of the
  International Conference on Learning Representations}, 2015.

\bibitem{sun2007adaptive}
Y.~Sun, Z.~Liu, S.~Todorovic, and J.~Li, ``Adaptive boosting for sar automatic
  target recognition,'' \emph{IEEE Transactions on Aerospace and Electronic
  Systems}, vol.~43, no.~1, pp. 112--125, 2007.

\bibitem{SVA}
B.~Peng, B.~Peng, J.~Zhou, J.~Xia, and L.~Liu, ``Speckle-variant attack: Toward
  transferable adversarial attack to sar target recognition,'' \emph{IEEE
  Geoscience and Remote Sensing Letters}, vol.~19, pp. 1--5, 2022.

\bibitem{smgaa}
B.~Peng, B.~Peng, J.~Zhou, J.~Xie, and L.~Liu, ``Scattering model guided
  adversarial examples for sar target recognition: Attack and defense,''
  \emph{IEEE Transactions on Geoscience and Remote Sensing}, vol.~60, pp.
  1--17, 2022.

\end{thebibliography}
\end{document}